\documentclass{article}

\usepackage[preprint]{neurips_2026}

\usepackage[utf8]{inputenc} 
\usepackage[T1]{fontenc}    
\usepackage{url}            
\usepackage{booktabs}       
\usepackage{amsfonts}       
\usepackage{nicefrac}       
\usepackage{microtype}      
\usepackage{xcolor}         

\usepackage{graphicx}
\usepackage{amsmath}
\usepackage{amssymb}
\usepackage{soul}

\usepackage{diagbox}
\usepackage{multicol}
\usepackage{enumerate}
\usepackage{times}
\usepackage{epsfig}
\usepackage{threeparttable}
\usepackage{enumitem}
\usepackage{multirow}
\usepackage{color}
\usepackage{array}
\usepackage{setspace}
\usepackage{makecell}
\usepackage{indentfirst}
\usepackage{wrapfig}

\definecolor{neuripsblue}{rgb}{0.21,0.49,0.74}
\usepackage[colorlinks, linkcolor=red, citecolor=neuripsblue]{hyperref}
\newcommand{\gh}[1]{{\color{black} {#1}}}

\title{TriRelVLA: Triadic Relational Structure for Generalizable Embodied Manipulation}

\author{Hanyu Zhou\textsuperscript{\rm 1}, Chuanhao Ma\textsuperscript{\rm 2}, Gim Hee Lee\textsuperscript{\rm 1}\\
  \textsuperscript{\rm 1} School of Computing, National University of Singapore\\
  \textsuperscript{\rm 2} School of Artificial Intelligence and Automation, Huazhong University of Science and Technology\\
  {\tt\small {\{hy.zhou, gimhee.lee\}}@nus.edu.sg}
 }

\begin{document}
\maketitle

\begin{abstract}
\gh{
Vision-language-action (VLA) models perform well on training-seen robotic tasks but struggle to generalize to unseen scenes and objects. A key limitation lies in their implicit visual representations, which entangle object appearance, background, and scene layout. This makes policies sensitive to visual variations. Prior work improves transferability through structured intermediate representations that objectify visual content. However, these representations mainly capture scene semantics instead of action-relevant relations. As a result, action prediction remains tied to appearance statistics. We observe that manipulation actions depend on the object–hand–task relational structure, which governs interactions among task requirements, robot states, and object properties. Based on this observation, we propose \textbf{TriRelVLA}, a triadic relational VLA framework for generalizable embodied manipulation. Our approach consists of three components: 1) We construct explicit \textbf{object–hand–task triadic representations} from multimodal inputs as relational primitives. 2) We build a \textbf{task-grounded relational graph}. Task-guided cross-attention forms nodes, and a relation-aware graph transformer models interactions among them. 3) We perform \textbf{relation-conditioned action generation}. The relational structure is compressed into a bottleneck space and projected into the LLM for action prediction. This triadic relational bottleneck reduces reliance on appearance statistics and enables transfer across scenes, objects, and task compositions. We further introduce a real-world robotic dataset for fine-tuning. Experiments show strong performance on fine-tuned tasks and clear gains in cross-scene, cross-object, and cross-task generalization.
}
\end{abstract}

\vspace{-2mm}
\section{Introduction}
\begin{figure}
  \setlength{\abovecaptionskip}{2pt}
  \setlength{\belowcaptionskip}{-5pt}
  \centering
   \includegraphics[width=0.99\linewidth]{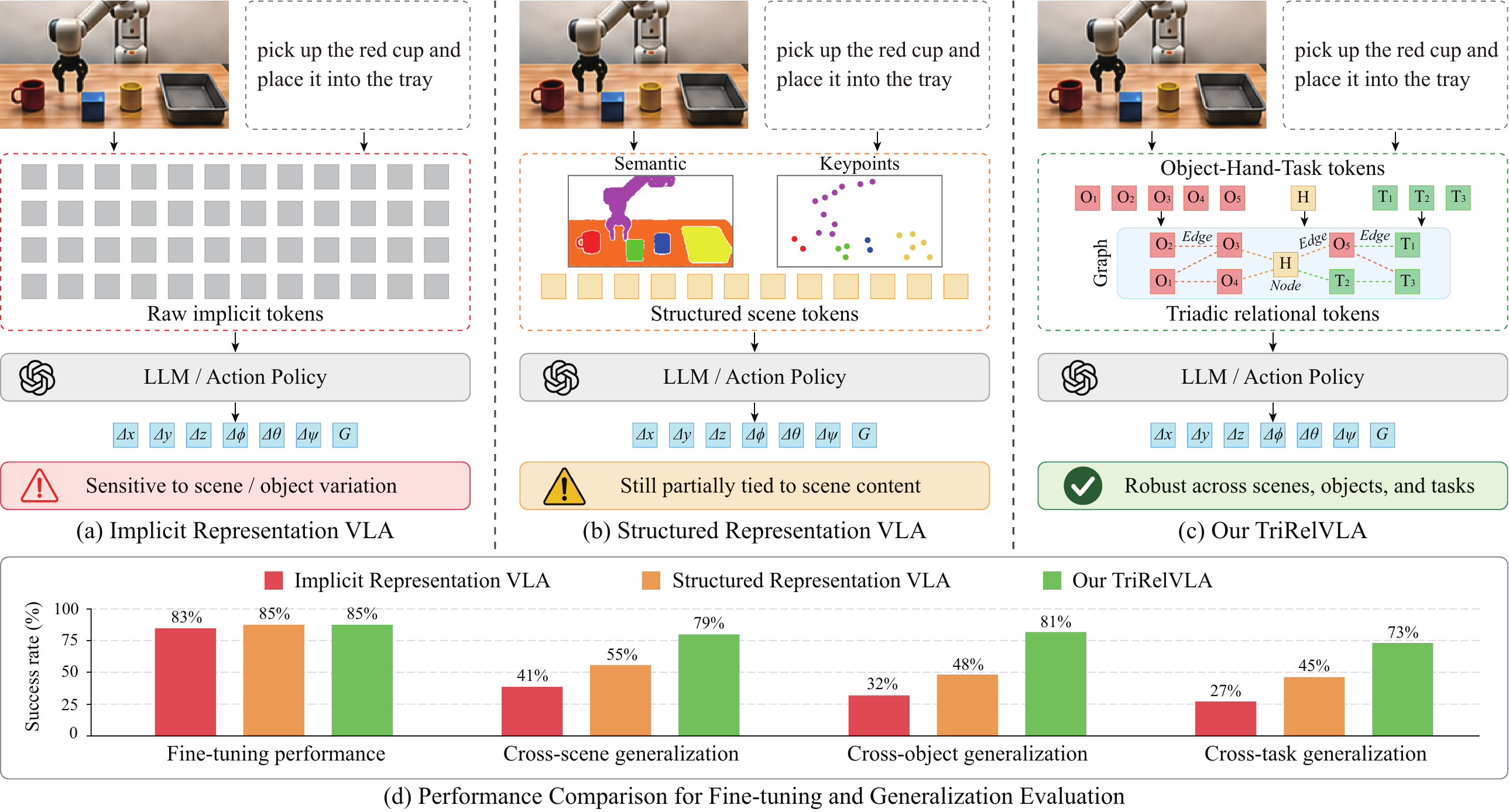}
   \caption{
   Illustration of VLA representation paradigms for robotic manipulation. (a) Implicit VLAs use dense latent tokens but are sensitive to scene variations. (b) Structured VLAs objectify visual content yet remain partially tied to scene appearance. (c) TriRelVLA models the object-hand-task relational structure for compact and transferable action prediction. (d) Results show competitive fine-tuning performance and stronger cross-scene, cross-object, and cross-task generalization.
   } 
   \label{Fig:Paradigm}
\end{figure}

\vspace{-2mm}
\label{sec:intro}
\gh{Vision-language-action (VLA) models have been widely used in embodied tasks such as manipulation \cite{brohan2022rt, kim2024openvla, kim2025fine} and navigation \cite{liu2024volumetric, wang2024vision}. As shown in Fig.~\ref{Fig:Paradigm}(a), existing VLAs \cite{kim2024openvla, black2024pi_0, zitkovich2023rt, chi2025diffusion, intelligence2025pi_, bjorck2025gr00t} encode visual observations and language instructions with LLMs to predict actions. They learn strong cross-modal action priors from large-scale demonstrations. However, their implicit visual representations entangle object appearance, background texture, and scene layout, which makes policies sensitive to appearance and scene variations. As a result, generalization to unseen scenes and objects remains limited as reflected in Fig.~\ref{Fig:Paradigm}(d).}
\gh{To address this issue, prior methods \cite{liu2024moka, huang2026graphcot, li2025object, li2024manipllm} improve transferability between perception and control by constructing intermediate representations that objectify visual content as summarized in Fig.~\ref{Fig:Paradigm}(b). However, these representations mainly capture scene semantics instead of action-relevant relations. This leaves action prediction partially tied to visual appearance statistics. In contrast, robotic manipulation requires reasoning over task requirements, robot states, object attributes, and their interactions. Explicit modeling of the relations among these cues is thus critical for cross-task generalization in VLAs.}

\gh{Motivated by this gap, we introduce the object–hand–task relational structure for action decisions in manipulation as illustrated in Fig.~\ref{Fig:Paradigm}(c). Specifically, we observe that action-relevant visual information centers on the target object and the robot hand, and language specifies task goals and constraints. We therefore propose a triadic representation of object, hand, and task as a bridge between visual perception and action decision, and model their interactions to drive action generation and filter redundant scene factors. This representation reduces dependence on appearance statistics and supports more compact and transferable action prediction.}


\gh{Building on the need for explicit relational modeling, we introduce \textbf{TriRelVLA}, a VLA framework with a triadic relational structure for generalizable embodied manipulation. As illustrated in Fig.~\ref{Fig:Framework}, the framework contains three components.
First, we construct an object–hand–task representation. Multimodal inputs are encoded into visual and language latent spaces. Object and hand representations are derived from visual features, and task representations from language features. These form the basic relational primitives.
Second, we build a task-grounded relational graph. Task-guided cross-attention organizes object and hand representations into nodes together with task representations. Pairwise edges define the relational structure, which is updated with a relation-aware graph transformer.
Third, we perform relation-conditioned action generation. The updated relational structure is compressed into a bottleneck space and projected into the LLM for action prediction.}

\gh{Our TriRelVLA is a unified design that directly addresses the limitations of prior representations, which improves data efficiency and compositional generalization across diverse scenes, objects, and task combinations. We also introduce a real-world robotic dataset to fine-tune the model for generalizable manipulation.}
Our main contributions are summarized as follows:
\begin{itemize}[leftmargin=*]
\item We propose a novel triadic relational VLA framework for generalizable embodied manipulation. 
\gh{Our model improves the transferability of VLAs to unseen robotic scenes with a triadic relational structure as an intermediate representation between visual perception and action control.}

\item 
\gh{We design an explicit triadic representation that reveals the core cues governing action decision, and motivates an object–hand–task structure to organize visual representations and provide a more compact basis for action prediction.}

\item 
\gh{We construct a task-grounded relational graph by observing a strong interaction between triadic representations and action generation, which motivates a task-guided graph to optimize the relational structure and provide transferable support for action prediction.}

\item 
\gh{We build a real-world robotic dataset to fine-tune our model, and extensive experiments show that the framework achieves state-of-the-art performance across diverse generalizable robotic tasks.}
\end{itemize}

\vspace{-2mm}
\section{Related Work}\vspace{-2mm}
\label{sec:related}
\textbf{Vision-Language-Action Models.}
Vision-language models \cite{liu2023llava, alayrac2022flamingo, liu2023improved, li2023blip2, zhou2025llava} have advanced scene reasoning and inspired VLA models for embodied manipulation. 
Existing VLAs \cite{kim2024openvla, black2024pi_0, team2024octo, zhou2025vla, ma2026st, wen2025dexvla, fan2025long, zhou2025chatvla} typically align multi-view visual latents with language embeddings, and feed them into an LLM to learn action generation from large-scale demonstrations. 
However, their implicit visual representations entangle object appearance, background texture, and scene layout, making policies sensitive to unseen scene and object variations. 
In this work, we learn discriminative and structured scene representations for generalizable embodied manipulation.

\textbf{Generalizable Structured Representation.}
Improving the generalization of VLA models \cite{zheng2024tracevla, kim2025fine, wen2025tinyvla, zhao2025cot, huang2025graphcot} to unseen scenes and objects requires extracting structured cues that remain stable across environmental variations. 
Existing methods \cite{liu2024moka, huang2026graphcot, li2025object, li2024manipllm} construct intermediate representations, such as objects, regions, semantic maps, or keypoints, to objectify visual observations and improve transferability between perception and control. 
However, these representations mainly describe scene semantics rather than the action-relevant relations that govern manipulation decisions. As a result, action prediction remains partially tied to visual appearance statistics, limiting generalization across scenes, objects, and tasks. This motivates us to build an action-centric relational representation that explicitly structures the core cues required for compositional manipulation.

\textbf{Relational Reasoning for Action Generation.}
Mining task-relevant cues and modeling their relational structure is crucial for improving compositional generalization in visual reasoning and embodied decision. 
Existing relational modeling methods \cite{vaswani2017attention, ma2022relvit, xie2025relationlmm, huang2024structure, huang2024rekep, wang20254d, huang2026graphcot} can be broadly grouped into implicit attention-based reasoning and explicit structure-based reasoning. 
The former \cite{ma2022relvit, kim2024openvla, zhou2025vla} captures interactions through Transformer self-attention, but lacks explicit structural inductive bias for stable decision cue extraction. 
The latter \cite{huang2024structure, huang2024rekep, huang2026graphcot} constructs relations over intermediate entities such as objects, regions, keypoints, or scene graphs, but primarily serves scene understanding rather than action generation. 
In contrast, TriRelVLA explicitly models task-grounded object-hand-task relations, forming a compact action-centric relational representation for compositional generalization across unseen scenes, objects, and tasks.

\vspace{-2mm}
\section{Our TriRelVLA}
\begin{figure}
  \setlength{\abovecaptionskip}{0pt}
  \setlength{\belowcaptionskip}{-7pt}
  \centering
   \includegraphics[width=0.99\linewidth]{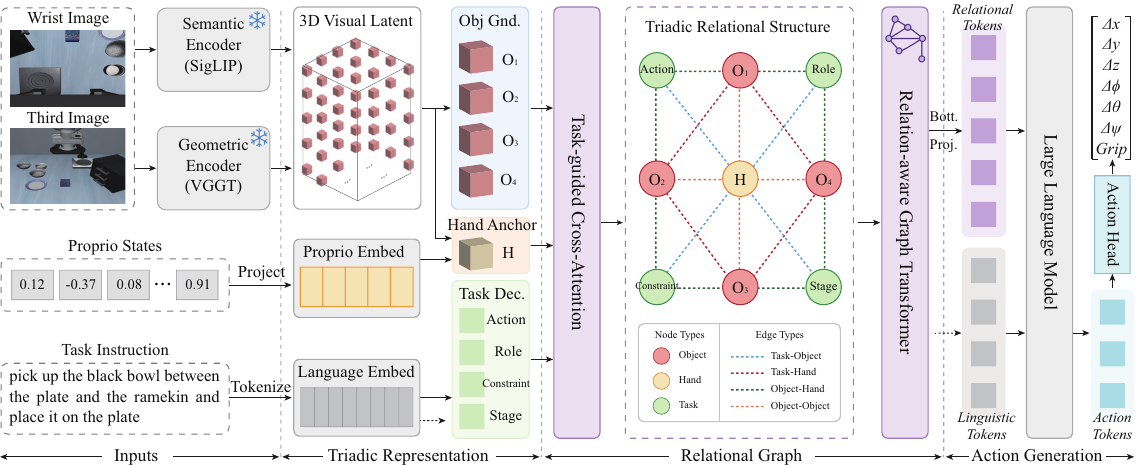}
   \caption{
   Our \textbf{TriRelVLA} consists of three stages: \textbf{1) Triadic representation.} Multi-view images, proprioceptive states, and task instructions are encoded to construct object, hand, and task tokens. \textbf{2) Relational graph.} The triadic tokens are transformed into task-grounded nodes and updated by a relation-aware graph transformer. \textbf{3) Action generation.} The relation-enhanced nodes are compressed into a bottleneck space and fed into the LLM with linguistic tokens for action prediction.
   }\vspace{-2mm}
   \label{Fig:Framework}
\end{figure}

\vspace{-2mm}
\label{sec:method}
\textbf{Overview.}
Fig.~\ref{Fig:Framework} shows the architecture of our TriRelVLA. Given multi-view images and instruction texts, our TriRelVLA achieves generalizable embodied manipulation through the three stages:
\vspace{-2mm}
\begin{enumerate}[leftmargin=*, label=\arabic*)]
    \item \textbf{Object-Hand-Task Triadic Representation (
    Sec.~\ref{sec:triadic})}. 
    \gh{This stage constructs a triadic representation from visual and linguistic inputs. We first encode the image and text into a visual latent $\mathbf{F}_v$ and a linguistic latent $\mathbf{F}_l$. We then derive object, hand, and task tokens with dedicated operators:}
    \begin{equation} 
    \setlength\abovedisplayskip{2mm}
    \setlength\belowdisplayskip{2mm}
        \begin{aligned}
            \mathbf{Z}_o = \operatorname{Ground}(\mathbf{F}_v), \quad
            \mathbf{Z}_h = \operatorname{Anchor}(\mathbf{F}_v, \mathbf{F}_p), \quad
            \mathbf{Z}_t = \operatorname{Decomp}(\mathbf{F}_l),
        \end{aligned}
        \label{eq:triadic_representation}
    \end{equation}
    \gh{where $\mathbf{Z}_o$, $\mathbf{Z}_h$, and $\mathbf{Z}_t$ denote the object, hand, and task tokens that form the triadic structural primitive. Here, $\mathbf{F}_p$ is the robot proprioceptive embedding used to condition hand token extraction.}
    \item \textbf{Task-Grounded Relational Graph (
    Sec.~\ref{sec:graph})}. This stage models the triadic relational structure. We transform the triadic tokens into nodes $\mathcal{V}$ through task-guided cross-attention, and define pairwise edges $\mathcal{E}$ to form a relational structure, which is updated by a graph transformer:
    \begin{equation} 
    \setlength\abovedisplayskip{2mm}
    \setlength\belowdisplayskip{2mm}
        \begin{aligned}
            \mathcal{V} = \operatorname{CAtt}([\mathbf{Z}_o \parallel \mathbf{Z}_h \parallel \mathbf{Z}_t]), \quad
            \mathcal{E} = \{ \mathbf{e}_{ij} \mid \mathbf{v}_i, \mathbf{v}_j \in \mathcal{V}, i \neq j \}, \quad
            \hat{\mathcal{V}} = \operatorname{GraphTrans}(\mathcal{V}, \mathcal{E}),
        \end{aligned}
        \label{eq:relational_graph}
    \end{equation}
    where $\hat{\mathcal{V}}$ denotes the relation-enhanced triadic nodes.
    \item \textbf{Relation-Conditioned Action Generation (
    Sec.~\ref{sec:action})}. This stage designs conditional action generation. We compress the updated triadic nodes into a bottleneck space: $\mathbf{R}=\operatorname{Bott}(\hat{\mathcal{V}})$, and then project them into the language embedding space: $\mathbf{X}_r = \operatorname{Proj}(\mathbf{R})$, where $\mathbf{X}_r$ denotes the relational tokens aligned with the linguistic tokens $\mathbf{X}_l$. The LLM then learns the mapping from the aligned multimodal tokens to actions: $\hat{\mathbf{a}}=\operatorname{LLM}([\mathbf{X}_l \parallel \mathbf{X}_r])$.
\end{enumerate}

\textbf{Remarks.} The output $\hat{\mathbf{a}}$ denotes the final action parameters determined by the triadic relational structure. 
Our framework uses triadic relations as an intermediate bridge between perception and control, improving data efficiency and compositional generalization across scenes, objects, and tasks.

\vspace{-2mm}
\subsection{Object-Hand-Task Triadic Representation}
\label{sec:triadic}
\vspace{-2mm}
Existing VLA models typically learn an implicit mapping from images and instructions to actions. 
\gh{However, this strategy introduces distractions that hinder generalization due to the conditioning of action generation on high-dimensional visual features entangled with irrelevant scene factors.}
In this section, we extract action-relevant cues to improve robustness against redundant information.

\textbf{Multimodal Embedding.}
In VLA feature space, visual and action representations often lie in different coordinates: images are in 2D pixel space, while actions are executed in 3D world or robot coordinates. 
To reduce this mismatch, we construct a unified 3D visual latent. Specifically, we use SigLIP \cite{zhai2023sigmoid} to extract multi-view semantic features $\mathbf{F}_{sem}$, and VGGT \cite{wang2025vggt} to obtain 3D geometric features $\mathbf{F}_{geo}$. These features are jointly fused into a unified 3D visual latent $\mathbf{F}_v$ for both wrist-view and third-view observations 
\gh{to facilitate} more consistent visual-to-action feature mapping.
For language and proprioception, we tokenize the instruction text into the linguistic latent $\mathbf{F}_l$, and optionally project the robot proprioceptive state into a learnable embedding space: $\mathbf{F}_p = \operatorname{Proj}_p(\mathbf{p})$, where $\mathbf{p}$ denotes the raw proprioceptive parameters.

\begin{wrapfigure}{r}{0.62\columnwidth} 
  \setlength{\abovecaptionskip}{3pt}
  \setlength{\belowcaptionskip}{-6pt}
  \vspace{-10pt} 
  \centering
   \includegraphics[width=0.62\columnwidth]{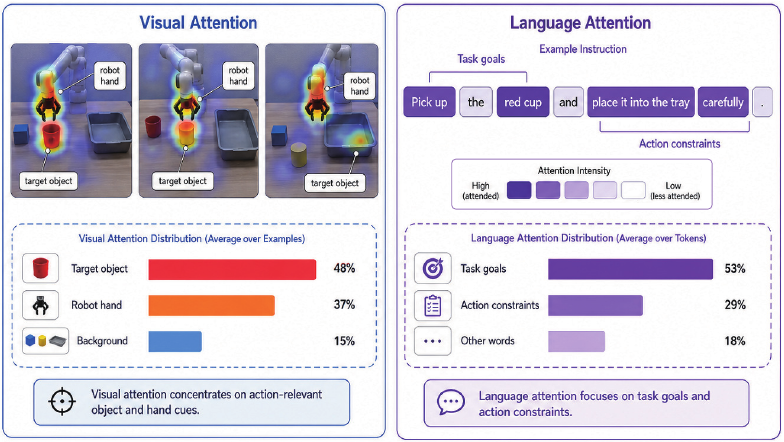}
   \caption{
   Visualization of action-relevant attention patterns.
   } 
   \label{Fig:Representation}
\end{wrapfigure}

\textbf{Triadic Representation.}
The visual latent contains background content and distractors that may obscure action-relevant cues. 
We 
\gh{posit} that action prediction should depend on task requirements, robot states, object properties, and their interactions 
\gh{instead of} holistic scene recognition. 
To identify these cues, we extract visual and language embeddings from a pretrained VLA model such as OpenVLA \cite{kim2024openvla} and visualize the attention distributions for action prediction. 
As shown in Fig.~\ref{Fig:Representation}, visual attention mainly focuses on target objects and the robot hand, 
\gh{and} language attention highlights task goals and action constraints. 
\gh{We are motivated by this observation to} use object, hand, and task representations as primitives for action decision. 
\gh{
For \textit{\textbf{object representation}}, we ground candidate object tokens from the 3D visual latent using learnable object queries $\mathbf{Q}_o$. These queries attend to visual features $\mathbf{F}_v$ to aggregate object-centric evidence. Specifically, we compute projected queries, keys, and values as:
\begin{equation}
\setlength\abovedisplayskip{1mm}
\setlength\belowdisplayskip{1mm}
\tilde{\mathbf{Q}}_o = \mathbf{Q}_o \mathbf{W}_q^o,\quad
\tilde{\mathbf{K}}_o = \mathbf{F}_v \mathbf{W}_k^o,\quad
\tilde{\mathbf{V}}_o = \mathbf{F}_v \mathbf{W}_v^o,\quad
\end{equation}
followed by attention:
$\mathbf{Z}_o = \operatorname{Softmax}\left(\tilde{\mathbf{Q}}_o \tilde{\mathbf{K}}_o^\top / \sqrt{d}\right)\tilde{\mathbf{V}}_o$,
where $\mathbf{W}_q^o$, $\mathbf{W}_k^o$, and $\mathbf{W}_v^o$ are learnable projection matrices, and $\mathbf{Z}_o$ denotes the resulting candidate object tokens.
}
\gh{For \textbf{\textit{hand representation}}, we introduce a proprioception-anchored hand query to capture hand cues. The robot proprioceptive embedding $\mathbf{F}_p$ serves as the query, which attends to the 3D visual latent $\mathbf{F}_v$ to aggregate hand-centric evidence. Specifically, we compute:
\begin{equation}
\setlength\abovedisplayskip{1mm}
\setlength\belowdisplayskip{1mm}
    \tilde{\mathbf{Q}}_h = \mathbf{F}_p \mathbf{W}_q^h, \quad
    \tilde{\mathbf{K}}_h = \mathbf{F}_v \mathbf{W}_k^h, \quad
    \tilde{\mathbf{V}}_h = \mathbf{F}_v \mathbf{W}_v^h, \quad 
\end{equation}
followed by attention $\mathbf{Z}_h = \operatorname{Softmax} \left(\tilde{\mathbf{Q}}_h \tilde{\mathbf{K}}_h^\top / \sqrt{d}\right)\tilde{\mathbf{V}}_h$, where $\mathbf{W}_q^h$, $\mathbf{W}_k^h$, and $\mathbf{W}_v^h$ are learnable projection matrices, and $\mathbf{Z}_h$ denotes the resulting hand token.
}
\gh{For \textit{\textbf{task representation}}, we decompose the linguistic latent into four task-aware tokens: action, role, constraint, and stage. Four learnable category queries attend over $\mathbf{F}_l$ to aggregate distinct task cues. Specifically, we compute $\tilde{\mathbf{F}}_l = \mathbf{F}_l \mathbf{W}_t$, followed by attention $\boldsymbol{\alpha}^{m} = \operatorname{Softmax} \left(\mathbf{q}_t^{m} \tilde{\mathbf{F}}_l^{\top} / \sqrt{d_t}\right)$ and $\mathbf{z}_t^{m} = \boldsymbol{\alpha}^{m} \tilde{\mathbf{F}}_l$, where $m \in \{\text{act}, \text{role}, \text{con}, \text{stage}\}$. Here, $\mathbf{q}_t^{m}$ denotes the learnable query for the $m$-th task category, and $\mathbf{z}_t^{m}$ denotes the corresponding task token.}
The final task token is defined: $\mathbf{Z}_t = [\mathbf{z}_t^\text{act} \parallel \mathbf{z}_t^\text{role} \parallel \mathbf{z}_t^\text{con} \parallel \mathbf{z}_t^\text{stage}]$ as structured task primitives.

\vspace{-2mm}
\subsection{Task-Grounded Relational Graph}
\label{sec:graph}
\vspace{-2mm}
\gh{Despite structured visual representations, existing VLAs generalize poorly across scenes and objects due to underexplored relations between triadic representations and action decisions. We thus explicitly model triadic relations to support generalizable action generation.}

\begin{wrapfigure}{r}{0.62\columnwidth} 
  \setlength{\abovecaptionskip}{0pt}
  \setlength{\belowcaptionskip}{-5pt}
  \vspace{-10pt} 
  \centering
   \includegraphics[width=0.62\columnwidth]{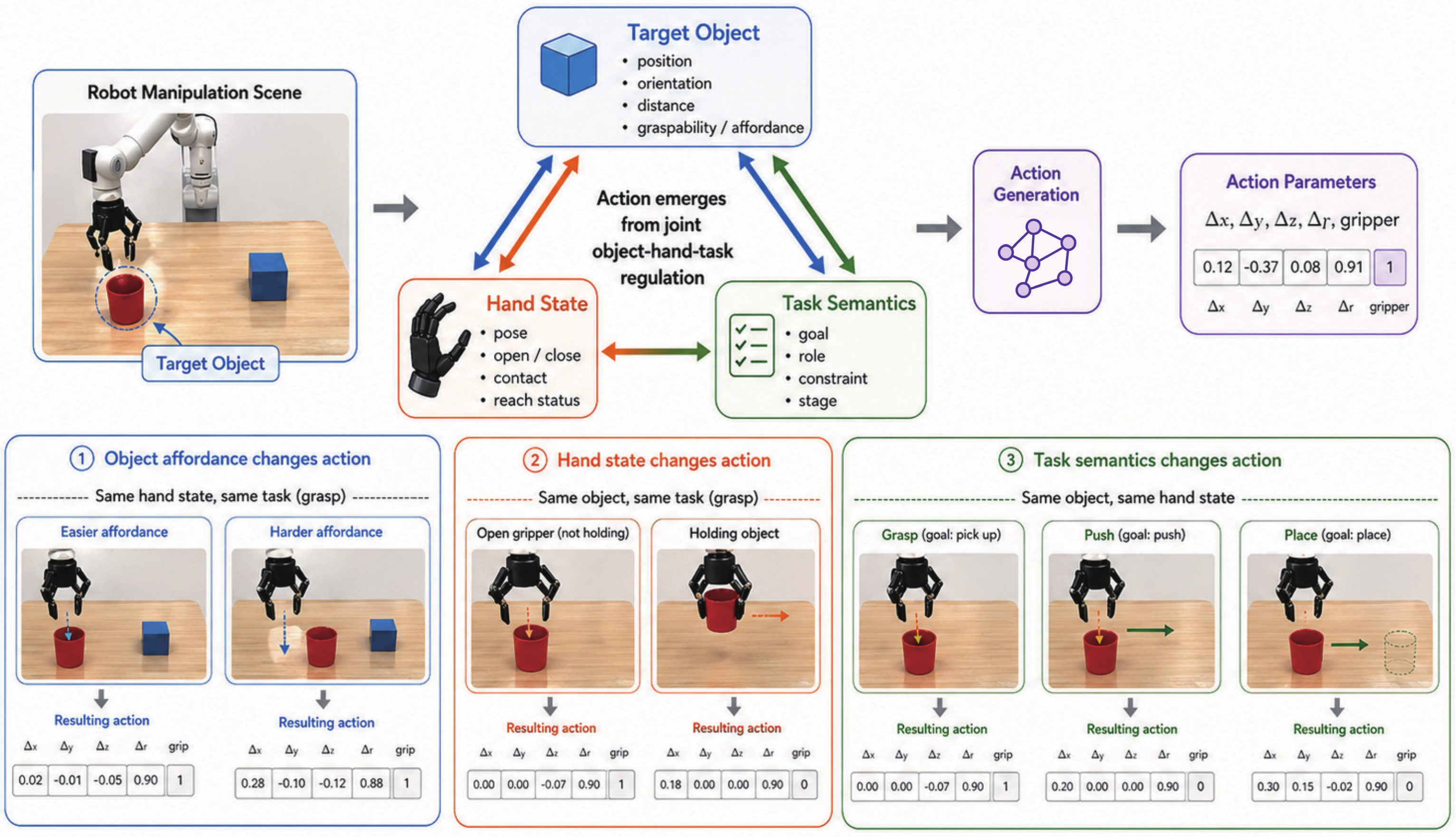}
   \caption{
   Illustration of triadic interaction for action decision.
   } 
   \label{Fig:Relation}
\end{wrapfigure}

\textbf{Task-Guided Triadic Nodes \& Edges.}
The relational structure among triadic representations is crucial for cross-scene and cross-task transfer. 
Scene graphs provide an explicit and interpretable way to model how key cues jointly govern action generation. 
As shown in Fig.~\ref{Fig:Relation}, action generation 
\gh{depends mainly} on target-object affordances, hand states, and task constraints 
\gh{to form} an object-hand-task interaction pattern. 
We therefore refine the scene graph into an object-hand-task relational structure to focus on action-relevant interactions while suppressing redundant scene factors. 
\gh{
For \textbf{\textit{node construction}}, we use task-guided cross-attention to inject task context into object and hand tokens to form graph nodes. Specifically, we first compute the projected queries, keys, and values:
\begin{equation}
\setlength\abovedisplayskip{2mm}
\setlength\belowdisplayskip{2mm}
\mathbf{Q}_o^n = \mathbf{Z}_o \mathbf{W}_{q,o}^n, \quad
\mathbf{Q}_h^n = \mathbf{Z}_h \mathbf{W}_{q,h}^n, \quad
\mathbf{K}_t^n = \mathbf{Z}_t \mathbf{W}_{k,t}^n, \quad
\mathbf{V}_t^n = \mathbf{Z}_t \mathbf{W}_{v,t}^n,
\end{equation}
followed by task-guided attention:
\begin{equation}\small
\setlength\abovedisplayskip{2mm}
\setlength\belowdisplayskip{2mm}
\tilde{\mathbf{Z}}_o = \mathbf{Z}_o + \operatorname{Softmax} \left(\frac{\mathbf{Q}_o^n (\mathbf{K}_t^n)^\top}{\sqrt{d}}\right)\mathbf{V}_t^n, \quad
\tilde{\mathbf{Z}}_h = \mathbf{Z}_h + \operatorname{Softmax} \left(\frac{\mathbf{Q}_h^n (\mathbf{K}_t^n)^\top}{\sqrt{d}}\right)\mathbf{V}_t^n,
\end{equation}
where $\mathbf{W}_{q,o}^n$, $\mathbf{W}_{q,h}^n$, $\mathbf{W}_{k,t}^n$, and $\mathbf{W}_{v,t}^n$ are learnable projection matrices. The resulting node set is defined as $\mathcal{V} = [\tilde{\mathbf{Z}}_o \parallel \tilde{\mathbf{Z}}_h \parallel \mathbf{Z}_t]$.}
\gh{For \textbf{\textit{edge construction}}, we model four relation types: task–object ($\mathbf{e}_{to}$), task–hand ($\mathbf{e}_{th}$), object–hand ($\mathbf{e}_{oh}$), and object–object ($\mathbf{e}_{oo}$). Each edge is computed via a relation-specific encoder applied to concatenated features. Concretely, we define $\mathbf{e}_{oh} = \phi_{oh}([\tilde{\mathbf{Z}}_{h}; \tilde{\mathbf{Z}}_{o}; \Delta \mathbf{p}_{oh}; \Delta \mathbf{R}_{oh}; d_{oh}; s_{oh}])$ and $\mathbf{e}_{oo} = \phi_{oo}([\tilde{\mathbf{Z}}_{o,i}; \tilde{\mathbf{Z}}_{o,j}; \Delta \mathbf{p}_{i,j}; \Delta \mathbf{R}_{i,j}; d_{i,j}])$ to capture geometric and interaction cues, and $\mathbf{e}_{to} = \phi_{to} ([\mathbf{Z}_{t}; \tilde{\mathbf{Z}}_{o}])$ and $\mathbf{e}_{th} = \phi_{th}([\mathbf{Z}_{t}; \tilde{\mathbf{Z}}_{h}])$ to encode task-conditioned dependencies. Here, $\phi_{oh}$, $\phi_{oo}$, $\phi_{to}$, and $\phi_{th}$ denote relation-specific edge encoders, and $\Delta \mathbf{p}$, $\Delta \mathbf{R}$, $d$, and $s$ represent relative position, relative pose, distance, and contact or reachability score.
}
The resulting edge set is defined as: $\mathcal{E} = \{\mathbf{e}_{to}, \mathbf{e}_{th}, \mathbf{e}_{oh}, \mathbf{e}_{oo}\}$.

\textbf{Relation-Aware Graph Transformer.}
\gh{
To learn the relational structure among triadic nodes and edges, we design a relation-aware graph transformer that injects edge features directly into attention for interactions to be modeled and propagated across nodes. For each node $i$, we first compute projected queries, keys, values, and edge embeddings:
\begin{equation}
\setlength\abovedisplayskip{2mm}
\setlength\belowdisplayskip{2mm}
\mathbf{q}_i = \mathbf{W}_q^g \mathbf{v}_i, \quad
\mathbf{k}_j = \mathbf{W}_k^g \mathbf{v}_j,\quad
\mathbf{v}_j^{g} = \mathbf{W}_v^g \mathbf{v}_j, \quad
\mathbf{r}_{ij} = \mathbf{W}_r^g \mathbf{e}_{ij},
\end{equation}
followed by relation-aware attention and aggregation:
\begin{equation}
\small
\setlength\abovedisplayskip{2mm}
\setlength\belowdisplayskip{2mm}
\alpha_{ij} = \operatorname{Softmax}_{j}
\left(\frac{\mathbf{q}_i(\mathbf{k}_j+\mathbf{r}_{ij})^\top}{\sqrt{d}}\right), \quad
\hat{\mathbf{v}}_i = \mathbf{v}_i + \sum\nolimits_{j \in \mathcal{N}(i)}
\alpha_{ij}(\mathbf{v}_j^{g}+\mathbf{r}_{ij}).
\end{equation}
$\mathbf{v}_i$ denotes the feature of node $i$, $\mathbf{e}_{ij}$ the edge feature, and $\hat{\mathbf{v}}_i$ the relation-enhanced node feature. By incorporating $\mathbf{r}_{ij}$ into both attention weights and value updates, the model captures interactions that depend on node similarity as well as geometric and semantic relations. The final node set $\hat{\mathcal{V}} = \{\hat{\mathbf{v}}_i\}_{i=1}^{|\mathcal{V}|}$ provides structured and transferable representations for action prediction.
}

\vspace{-2mm}
\subsection{Relation-Conditioned Action Generation}
\label{sec:action}
\vspace{-2mm}
Although triadic relational representations capture object-hand-task interactions, directly feeding them into the LLM may cause weak cross-modal alignment. We therefore compress and align them into relational conditioning embeddings to guide task- and scene-aware action generation.

\textbf{Relational Bottleneck.}
Although the relational graph captures object-hand-task dependencies, it may still include redundant components such as task-irrelevant object nodes. 
We introduce a relational bottleneck to compress relation-enhanced nodes into compact action-relevant tokens 
\gh{to improve} alignment efficiency and generalization. 
Specifically, we compute an importance weight $\alpha_{k,i}$ for each node $\hat{\mathbf{v}}_i$ and aggregate them into $K$ relational bottleneck tokens:
\begin{equation}
    \setlength\abovedisplayskip{2mm}
    \setlength\belowdisplayskip{2mm}
    \begin{aligned}
    \alpha_{k,i} = \operatorname{Softmax}_{i}
    \left(
    \mathbf{w}_{k}^{\top}\hat{\mathbf{v}}_i
    \right), \quad
    \mathbf{r}_{k} = \sum\nolimits_{i=1}^{|\hat{\mathcal{V}}|}
    \alpha_{k,i}\hat{\mathbf{v}}_i, \quad
    \mathbf{R} = [\mathbf{r}_{1} \parallel \mathbf{r}_{2} \parallel \cdots \parallel \mathbf{r}_{K}],
    \end{aligned}
    \label{eq:relational_bottleneck}
\end{equation}
where $\mathbf{R}$ denotes the relational bottleneck representation, which helps further filter out irrelevant object interference and improves the data efficiency of the model.

\textbf{Conditional Action Generation.}
After obtaining the relational bottleneck, we project it into the language embedding space with an MLP: $\mathbf{X}_r=\operatorname{MLP}(\mathbf{R})$, where $\mathbf{X}_r$ denotes relational tokens. 
We then concatenate $\mathbf{X}_r$ with linguistic tokens $\mathbf{X}_l$ and feed them into the LLM for action prediction: $\hat{\mathbf{a}}=\mathcal{H}(\operatorname{LLM}([\mathbf{X}_l \parallel \mathbf{X}_r]))$, where $\mathcal{H}$ is the action head. 
\gh{Generalization and stability are improved by the reliance of action prediction on the compact triadic relations instead of raw visual appearance.}

\vspace{-2mm}
\subsection{Optimization}
\label{sec:optmization}
\vspace{-2mm}
To stabilize triadic representation learning and prevent token drift, we add object grounding and hand anchoring constraints besides action supervision. First, we employ L1 loss to supervise 
\gh{the prediction of continuous action parameters}: $\mathcal{L}_{act} = ||\hat{a}-a||_1$, where $\hat{a}$ and $a$ denote the predicted and ground-truth actions, respectively. 
Since object and hand representations are latent tokens, we supervise their attention maps 
\gh{instead of} token values by aligning them with the corresponding region masks: $\mathcal{L}_{obj} = \operatorname{BCE}(\mathbf{A}_o, \mathbf{M}_o)$, $\mathcal{L}_{hand} = \operatorname{BCE}(\mathbf{A}_h, \mathbf{M}_h)$, where $\mathbf{A}_o$, $\mathbf{A}_h$ denote the attention maps for \gh{the} object and hand, respectively. $\mathbf{M}_o$ is the object region mask and $\mathbf{M}_h$ is the hand region mask.
Note that the losses $\mathcal{L}_{obj}$ and $\mathcal{L}_{hand}$ are also applicable to the attention constraints of the object and hand nodes in the relational graph. The overall optimization objective is formulated as: $\mathcal{L}_{total} = \mathcal{L}_{act} + \lambda_o\mathcal{L}_{obj} + \lambda_h\mathcal{L}_{hand}$, where $\lambda_o$, $\lambda_h$ are the weighting coefficients for the auxiliary constraints. 
\gh{The} auxiliary losses stabilize object-hand alignment, 
\gh{and} action loss drives relational representations toward final action decisions.

\vspace{-2mm}
\section{Dataset and Training Pipeline}
\label{sec:dataset}
\vspace{-2mm}
\subsection{Robotic Dataset}
\vspace{-2mm}
To improve the basic action prediction ability, real-world transferability, and triadic relational modeling capability of our model, we adopt multiple complementary datasets to train our model.

\textbf{OXE.} Open X-Embodiment \cite{o2024open} is a large-scale robotic manipulation dataset for pre-training, covering diverse robots, environments, tasks, and basic skills such as grasping, placing, moving, opening/closing, pushing, and pulling.

\textbf{DROID.}
DROID \cite{khazatsky2024droid} is a large-scale real-world manipulation dataset collected in diverse household and tabletop scenes, emphasizing visual variation and manipulation complexity for improving real-scene transferability.
We further semi-automatically annotate DROID with hand and object masks.

\textbf{LIBERO.}
LIBERO \cite{liu2023libero} is a public simulation benchmark covering instruction-driven manipulation tasks in spatial reasoning, object understanding, goal completion, and long-horizon planning. We use it for generalization evaluation with object/hand masks for auxiliary supervision.

\begin{figure}
  \setlength{\abovecaptionskip}{0pt}
  \setlength{\belowcaptionskip}{-7pt}
  \centering
   \includegraphics[width=0.95\linewidth]{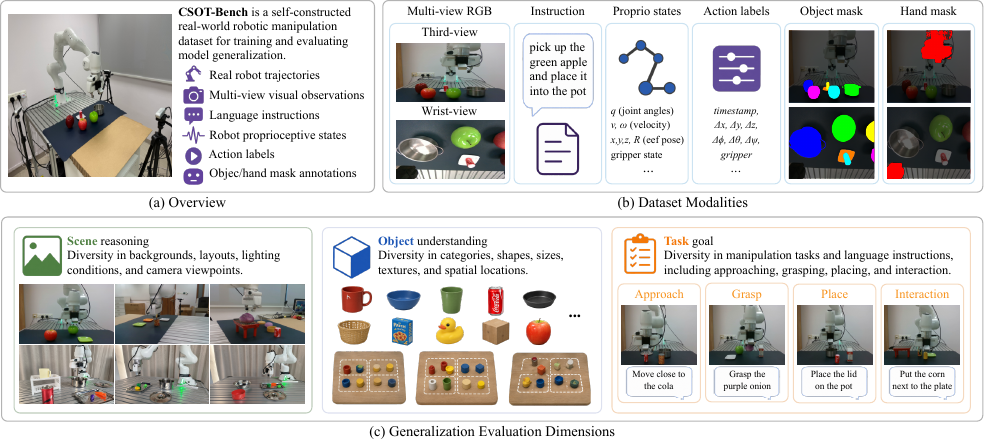}
   \caption{
   Illustration of CSOT-Bench. (a) Real-world robotic manipulation dataset. (b) Multi-modal annotations, including multi-view RGB images, instructions, proprioception, actions, and object/hand masks. (c) Generalization evaluation dimensions across scenes, objects, and tasks.
   } \vspace{-2mm}
   \label{Fig:Dataset}
\end{figure}

\textbf{CSOT-Bench.}
CSOT-Bench is our self-constructed real-world robotic manipulation dataset with three generalization evaluation suites, including scene reasoning, object understanding and task goal.
As shown in Fig.~\ref{Fig:Dataset}, it includes robot trajectories, multi-view observations, instructions, proprioceptive states, action labels, and object/hand masks. 
It varies scene factors such as backgrounds, and viewpoints; object factors such as categories, shapes, and locations; and task factors such as approaching, grasping, placing, and goal-conditioned interactions. 
These variations allow CSOT-Bench to evaluate whether models learn structured object-hand-task relations rather than visual appearance cues.

\vspace{-2mm}
\subsection{Training Pipeline}
\vspace{-2mm}
Our TriRelVLA model first loads the weights of several pretrained models to initialize its visual reasoning capability. To fully enhance the learning capability of the triadic relational structure for generalizable embodied tasks, we divide the training process into three stages:

\textbf{Stage 1: Multimodal Alignment.}
This stage learns basic vision-action alignment using the large-scale OXE dataset with only the action loss $\mathcal{L}_{act}$. 
It provides a general action prior for subsequent relational representation learning. 
At this stage, we only update the projector and the action head, while freezing the remaining modules.

\textbf{Stage 2: Relational Structure Modeling.}
This stage learns triadic representations and relational structures on the realistic DROID dataset with the full loss $\mathcal{L}_{total}$. 
It shifts the model from vision-action mapping to relation-driven action generation, improving generalization under complex scenes and object variations. 
At this stage, we update the triadic representation, relational graph, bottleneck space, and projector, while freezing the semantic encoder, geometric encoder, action head, and LLM.

\textbf{Stage 3: Robotic Task Fine-tuning.}
This stage fine-tunes the model on LIBERO and CSOT-Bench with the full loss $\mathcal{L}_{total}$ for task adaptation and compositional generalization. 
It adapts the model to specific tasks, objects, and scene variations while preserving prior vision-action alignment and relational modeling abilities. 
At this stage, we update the triadic representation, relational graph, bottleneck space, projector, and action head, and optimize the LLM components using LoRA, while keeping the semantic encoder and geometric encoder frozen.

\vspace{-2mm}
\section{Experiments}
\label{sec:experiments}
\vspace{-2mm}
\textbf{Implementation Details.}
TriRelVLA uses Qwen3-4B \cite{yang2025qwen3} as the LLM backbone, SigLIP \cite{zhai2023sigmoid} as the semantic encoder, and DINOv2 from VGGT \cite{wang2025vggt} as the geometry encoder. 
The model is trained on 8$\times$A800 GPUs with AdamW. 
The learning rates and batch sizes are set to $1\times10^{-4}$/64, $2\times10^{-5}$/32, and $1\times10^{-5}$/32 for stages 1, 2, and 3, respectively.

\textbf{Comparison Methods.}
We compare TriRelVLA with typical VLAs, including OpenVLA \cite{kim2024openvla}, Octo \cite{team2024octo}, CogACT \cite{li2024cogact}, DiffusionPolicy \cite{chi2025diffusion}, SpatialVLA \cite{qu2025spatialvla}, CoA-VLA \cite{li2024coa}, CogVLA \cite{li2025cogvla}, and SemanticVLA \cite{li2026semanticvla}. 
The first five methods directly learn implicit vision-to-action mappings, 
\gh{and} the latter three methods introduce intermediate representations to structure visual features.

\textbf{Evaluation Metric.}
We evaluate all methods on LIBERO and CSOT-Bench under two settings: fine-tuning on all subtasks and zero-shot generalization across scenes, objects, and tasks. 
Task success rate is used as the primary metric. 
Ablation and discussion experiments in Sec.~\ref{sec:ablation} are conducted on CSOT-Bench, with results reported for both fine-tuning and zero-shot generalization.

\begin{table*}[t]\footnotesize
    \setlength{\abovecaptionskip}{0pt}
    \setlength\tabcolsep{4pt}
    \setlength{\belowcaptionskip}{1pt}
    \caption{Quantitative results of VLAs for fine-tuned robotic manipulation tasks.}
  \centering
  \renewcommand\arraystretch{1.1}
  \begin{tabular}{ccccccccccc}
    \Xhline{1pt}
      \multicolumn{1}{c|}{\multirow{2}{*}{\makecell{Representation\\Category}}} &
      \multicolumn{1}{c|}{\multirow{2}{*}{Method}} &
      \multicolumn{5}{c|}{\multirow{1}{*}{LIBERO \cite{liu2023libero}}} &
      \multicolumn{4}{c}{\multirow{1}{*}{CSOT-Bench (Ours)}} \\      
      \cline{3-11}
      \multicolumn{1}{c|}{\multirow{1}{*}{}} &
      \multicolumn{1}{c|}{\multirow{1}{*}{}} &
      \multicolumn{1}{c}{\multirow{1}{*}{Spatial}} &
      \multicolumn{1}{c}{\multirow{1}{*}{Object}} &
      \multicolumn{1}{c}{\multirow{1}{*}{Goal}} &
      \multicolumn{1}{c}{\multirow{1}{*}{Long}} &
      \multicolumn{1}{c|}{\multirow{1}{*}{Average}} &
      \multicolumn{1}{c}{\multirow{1}{*}{Scene}} &
      \multicolumn{1}{c}{\multirow{1}{*}{Object}} &
      \multicolumn{1}{c}{\multirow{1}{*}{Task}} &
      \multicolumn{1}{c}{\multirow{1}{*}{Average}} \\
      \hline 

      \multicolumn{1}{c|}{\multirow{5}{*}{Implicit}} &
      \multicolumn{1}{c|}{\multirow{1}{*}{OpenVLA \cite{kim2024openvla}}} &
      \multicolumn{1}{c}{\multirow{1}{*}{84.7}} &
      \multicolumn{1}{c}{\multirow{1}{*}{88.4}} &
      \multicolumn{1}{c}{\multirow{1}{*}{79.2}} &
      \multicolumn{1}{c}{\multirow{1}{*}{53.7}} &
      \multicolumn{1}{c|}{\multirow{1}{*}{76.5}} &
      \multicolumn{1}{c}{\multirow{1}{*}{74.6}} &
      \multicolumn{1}{c}{\multirow{1}{*}{82.4}} &
      \multicolumn{1}{c}{\multirow{1}{*}{72.0}} &
      \multicolumn{1}{c}{\multirow{1}{*}{76.3}} \\
								
      \multicolumn{1}{c|}{\multirow{1}{*}{}} &
      \multicolumn{1}{c|}{\multirow{1}{*}{Octo \cite{team2024octo}}} &
      \multicolumn{1}{c}{\multirow{1}{*}{78.9}} &
      \multicolumn{1}{c}{\multirow{1}{*}{85.7}} &
      \multicolumn{1}{c}{\multirow{1}{*}{84.6}} &
      \multicolumn{1}{c}{\multirow{1}{*}{51.1}} &
      \multicolumn{1}{c|}{\multirow{1}{*}{75.1}} &
      \multicolumn{1}{c}{\multirow{1}{*}{66.3}} &
      \multicolumn{1}{c}{\multirow{1}{*}{78.0}} &
      \multicolumn{1}{c}{\multirow{1}{*}{71.5}} &
      \multicolumn{1}{c}{\multirow{1}{*}{72.0}} \\
 								
      \multicolumn{1}{c|}{\multirow{1}{*}{}} &
      \multicolumn{1}{c|}{\multirow{1}{*}{CogACT \cite{li2024cogact}}} &
      \multicolumn{1}{c}{\multirow{1}{*}{87.5}} &
      \multicolumn{1}{c}{\multirow{1}{*}{90.2}} &
      \multicolumn{1}{c}{\multirow{1}{*}{78.4}} &
      \multicolumn{1}{c}{\multirow{1}{*}{53.2}} &
      \multicolumn{1}{c|}{\multirow{1}{*}{76.5}} &
      \multicolumn{1}{c}{\multirow{1}{*}{76.1}} &
      \multicolumn{1}{c}{\multirow{1}{*}{84.5}} &
      \multicolumn{1}{c}{\multirow{1}{*}{73.8}} &
      \multicolumn{1}{c}{\multirow{1}{*}{78.1}} \\
								
      \multicolumn{1}{c|}{\multirow{1}{*}{}} &
      \multicolumn{1}{c|}{\multirow{1}{*}{DiffusionPolicy \cite{chi2025diffusion}}} &
      \multicolumn{1}{c}{\multirow{1}{*}{78.3}} &
      \multicolumn{1}{c}{\multirow{1}{*}{92.5}} &
      \multicolumn{1}{c}{\multirow{1}{*}{68.3}} &
      \multicolumn{1}{c}{\multirow{1}{*}{50.5}} &
      \multicolumn{1}{c|}{\multirow{1}{*}{72.4}} &
      \multicolumn{1}{c}{\multirow{1}{*}{68.7}} &
      \multicolumn{1}{c}{\multirow{1}{*}{80.2}} &
      \multicolumn{1}{c}{\multirow{1}{*}{65.3}} &
      \multicolumn{1}{c}{\multirow{1}{*}{71.4}} \\

      \multicolumn{1}{c|}{\multirow{1}{*}{}} &
      \multicolumn{1}{c|}{\multirow{1}{*}{SpatialVLA \cite{qu2025spatialvla}}} &
      \multicolumn{1}{c}{\multirow{1}{*}{88.2}} &
      \multicolumn{1}{c}{\multirow{1}{*}{89.9}} &
      \multicolumn{1}{c}{\multirow{1}{*}{78.6}} &
      \multicolumn{1}{c}{\multirow{1}{*}{55.5}} &
      \multicolumn{1}{c|}{\multirow{1}{*}{78.1}} &
      \multicolumn{1}{c}{\multirow{1}{*}{79.5}} &
      \multicolumn{1}{c}{\multirow{1}{*}{87.6}} &
      \multicolumn{1}{c}{\multirow{1}{*}{75.4}} &
      \multicolumn{1}{c}{\multirow{1}{*}{80.8}} \\
								
      \hline 

      \multicolumn{1}{c|}{\multirow{4}{*}{Structured}} &
      \multicolumn{1}{c|}{\multirow{1}{*}{CoA-VLA \cite{li2024coa}}} &
      \multicolumn{1}{c}{\multirow{1}{*}{85.3}} &
      \multicolumn{1}{c}{\multirow{1}{*}{93.1}} &
      \multicolumn{1}{c}{\multirow{1}{*}{85.8}} &
      \multicolumn{1}{c}{\multirow{1}{*}{55.0}} &
      \multicolumn{1}{c|}{\multirow{1}{*}{79.8}} &
      \multicolumn{1}{c}{\multirow{1}{*}{--}} &
      \multicolumn{1}{c}{\multirow{1}{*}{--}} &
      \multicolumn{1}{c}{\multirow{1}{*}{--}} &
      \multicolumn{1}{c}{\multirow{1}{*}{--}} \\
				
      \multicolumn{1}{c|}{\multirow{1}{*}{}} &
      \multicolumn{1}{c|}{\multirow{1}{*}{CogVLA \cite{li2025cogvla}}} &
      \multicolumn{1}{c}{\multirow{1}{*}{\textbf{98.6}}} &
      \multicolumn{1}{c}{\multirow{1}{*}{98.8}} &
      \multicolumn{1}{c}{\multirow{1}{*}{96.6}} &
      \multicolumn{1}{c}{\multirow{1}{*}{\textbf{95.4}}} &
      \multicolumn{1}{c|}{\multirow{1}{*}{97.4}} &
      \multicolumn{1}{c}{\multirow{1}{*}{81.2}} &
      \multicolumn{1}{c}{\multirow{1}{*}{88.1}} &
      \multicolumn{1}{c}{\multirow{1}{*}{78.4}} &
      \multicolumn{1}{c}{\multirow{1}{*}{82.5}} \\
								
      \multicolumn{1}{c|}{\multirow{1}{*}{}} &
      \multicolumn{1}{c|}{\multirow{1}{*}{SemanticVLA \cite{li2026semanticvla}}} &
      \multicolumn{1}{c}{\multirow{1}{*}{\textbf{98.6}}} &
      \multicolumn{1}{c}{\multirow{1}{*}{\textbf{99.6}}} &
      \multicolumn{1}{c}{\multirow{1}{*}{\underline{97.6}}} &
      \multicolumn{1}{c}{\multirow{1}{*}{\underline{94.8}}} &
      \multicolumn{1}{c|}{\multirow{1}{*}{\textbf{97.7}}} &
      \multicolumn{1}{c}{\multirow{1}{*}{\underline{82.7}}} &
      \multicolumn{1}{c}{\multirow{1}{*}{\underline{91.3}}} &
      \multicolumn{1}{c}{\multirow{1}{*}{\underline{78.5}}} &
      \multicolumn{1}{c}{\multirow{1}{*}{\underline{84.2}}} \\
								
      \multicolumn{1}{c|}{\multirow{1}{*}{}} &
      \multicolumn{1}{c|}{\multirow{1}{*}{TriRelVLA (Ours)}} &
      \multicolumn{1}{c}{\multirow{1}{*}{\underline{98.2}}} &
      \multicolumn{1}{c}{\multirow{1}{*}{\underline{99.0}}} &
      \multicolumn{1}{c}{\multirow{1}{*}{\textbf{97.8}}} &
      \multicolumn{1}{c}{\multirow{1}{*}{\underline{94.8}}} &
      \multicolumn{1}{c|}{\multirow{1}{*}{\underline{97.6}}} &
      \multicolumn{1}{c}{\multirow{1}{*}{\textbf{83.1}}} &
      \multicolumn{1}{c}{\multirow{1}{*}{\textbf{91.5}}} &
      \multicolumn{1}{c}{\multirow{1}{*}{\textbf{80.3}}} &
      \multicolumn{1}{c}{\multirow{1}{*}{\textbf{84.9}}} \\
      								
       \Xhline{1pt}
       
  \end{tabular} 
   \label{tab:comparison}\vspace{-2mm}
\end{table*}

\vspace{-2mm}
\subsection{Comparison with State-of-the-Art Models}
\label{sec:comparison}
\vspace{-2mm}
\textbf{Comparison on Fine-Tuned Tasks.}
Table~\ref{tab:comparison} compares fine-tuning performance on LIBERO and CSOT-Bench. TriRelVLA achieves performance comparable to existing VLAs, 
\gh{which indicates} that the triadic relational structure does not compromise basic manipulation accuracy. 
Although dense visual tokens are replaced by a compact relational bottleneck, the model still preserves sufficient task-relevant information for accurate action prediction.

\begin{figure}
  \setlength{\abovecaptionskip}{0pt}
  \setlength{\belowcaptionskip}{-7pt}
  \centering
   \includegraphics[width=0.99\linewidth]{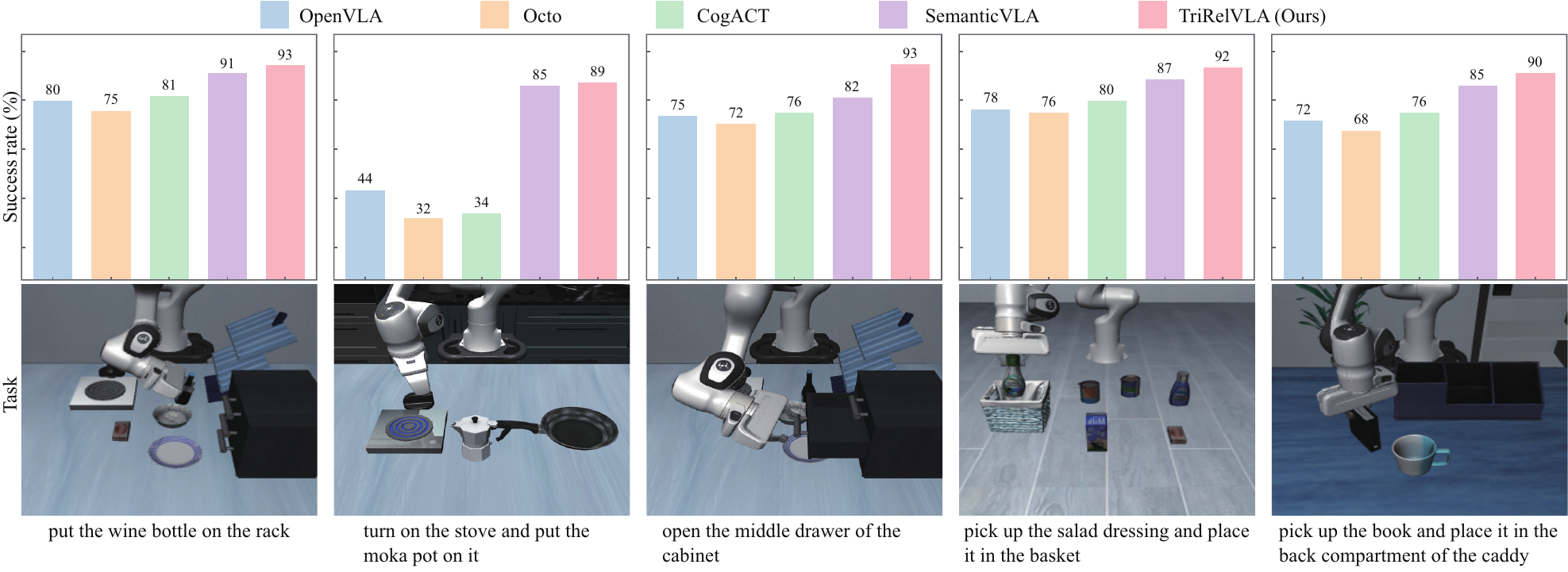}
   \caption{
   Quantitative comparison of VLAs for generalization evaluation on LIBERO.
   } \vspace{-2mm}
   \label{Fig:SynComparison}
\end{figure}

\begin{figure}[t]
  \setlength{\abovecaptionskip}{0pt}
  \setlength{\belowcaptionskip}{-7pt}
  \centering
   \includegraphics[width=0.99\linewidth]{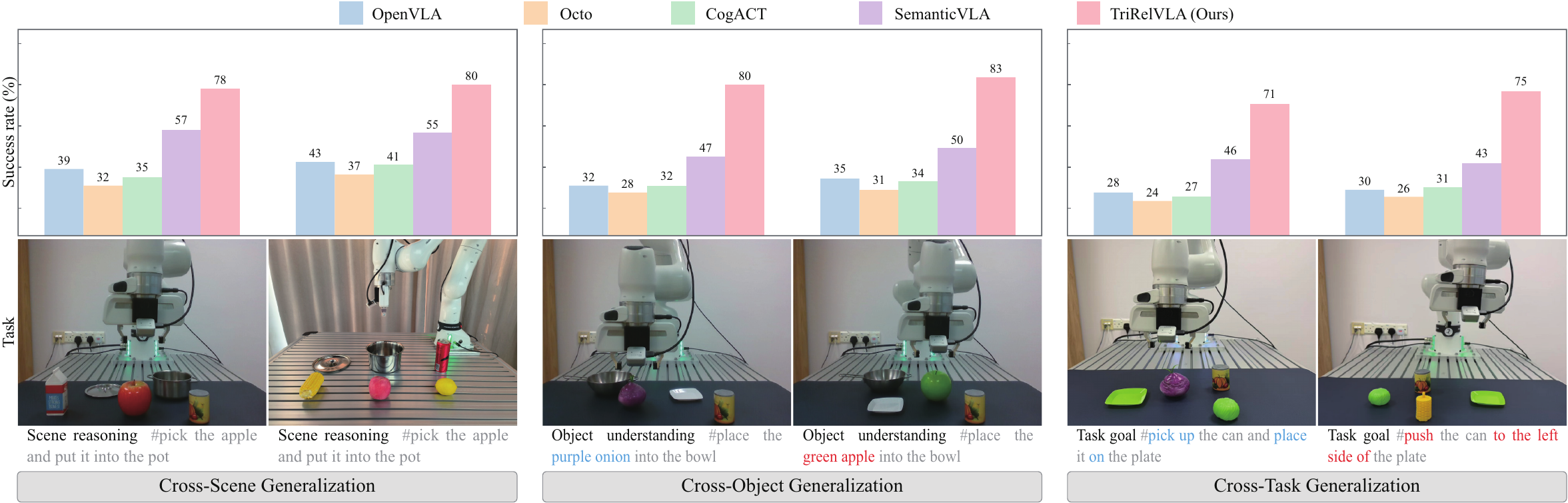}
   \caption{
   Generalization evaluation of VLAs on CSOT-Bench across scenes, objects, and tasks.
   } \vspace{-2mm}
   \label{Fig:RealComparison}
\end{figure}

\textbf{Comparison on Generalizable Robotic Tasks.}
Fig.~\ref{Fig:SynComparison} and Fig.~\ref{Fig:RealComparison} compare generalization across scenes, objects, and tasks on LIBERO and CSOT-Bench. 
Structured VLAs generally outperform implicit VLAs, while TriRelVLA achieves the best results, especially in cross-task generalization. 
The gains are modest in simulation but more pronounced in real-world settings, showing that object-hand-task relations reduce reliance on visual appearance and enable transferable action generation.

\vspace{-2mm}
\subsection{Ablation Study and Discussion}
\label{sec:ablation}
\vspace{-2mm}

\begin{wraptable}{r}{0.5\textwidth}\scriptsize
  \centering \vspace{-6mm}
  \caption{Effect of triadic relational architecture.}
  \setlength\tabcolsep{5pt}
  \renewcommand\arraystretch{1.0}
    \begin{tabular}{ccccc}
    \Xhline{1pt}
    \multicolumn{1}{c}{\multirow{2}{*}{\makecell{Basic\\arch.}}} & 
    \multicolumn{1}{c}{\multirow{2}{*}{\makecell{Triadic\\represent}}} & 
    \multicolumn{1}{c|}{\multirow{2}{*}{\makecell{Relational\\graph}}} & 
    \multicolumn{1}{c}{\multirow{2}{*}{\makecell{Fine-tuned\\task}}} & 
    \multicolumn{1}{c}{\multirow{2}{*}{\makecell{Generalized\\task}}} \\
    
    \multicolumn{1}{c}{\multirow{1}{*}{}} & 
    \multicolumn{1}{c}{\multirow{1}{*}{}} & 
    \multicolumn{1}{c|}{\multirow{1}{*}{}} \\
    
    \hline 
    
    \multicolumn{1}{c}{\multirow{1}{*}{$\surd$}} & 
    \multicolumn{1}{c}{\multirow{1}{*}{$\times$}} & 
    \multicolumn{1}{c|}{\multirow{1}{*}{$\times$}} & 
    \multicolumn{1}{c}{\multirow{1}{*}{82.6}} & 
    \multicolumn{1}{c}{\multirow{1}{*}{32.1}} \\

    \multicolumn{1}{c}{\multirow{1}{*}{$\surd$}} & 
    \multicolumn{1}{c}{\multirow{1}{*}{$\surd$}} & 
    \multicolumn{1}{c|}{\multirow{1}{*}{$\times$}} & 
    \multicolumn{1}{c}{\multirow{1}{*}{84.2}} & 
    \multicolumn{1}{c}{\multirow{1}{*}{43.0}} \\

    \multicolumn{1}{c}{\multirow{1}{*}{$\surd$}} & 
    \multicolumn{1}{c}{\multirow{1}{*}{$\surd$}} & 
    \multicolumn{1}{c|}{\multirow{1}{*}{$\surd$}} & 
    \multicolumn{1}{c}{\multirow{1}{*}{\textbf{84.9}}} & 
    \multicolumn{1}{c}{\multirow{1}{*}{\textbf{79.5}}} \\

    \Xhline{1pt}
       
  \end{tabular} 
  \vspace{-3mm}
  \label{Tab:Ablation_Architecture}
\end{wraptable}

\textbf{Effectiveness of Triadic Relational Architecture.}
Table~\ref{Tab:Ablation_Architecture} verifies the effectiveness of the basic architecture, triadic representations, and relational graph. The basic VLA maintains standard performance but shows limited generalization. Adding triadic representations improves generalization, while the relational graph brings further gains. 
This shows that object-hand-task modeling shifts the model from vision-to-action mapping to relation-driven action generation.

\begin{wraptable}{r}{0.45\textwidth}
\vspace{-2mm}
\centering
\scriptsize
\setlength{\tabcolsep}{6pt}
\renewcommand{\arraystretch}{0.95}
\caption{Impact of structured representations.}
\begin{tabular*}{0.45\textwidth}{@{\extracolsep{\fill}}ccc}
\Xhline{1pt}
\makecell{Representation strategy} & 
\makecell{Fine-tuned task} & 
\makecell{Generalized task} \\
\hline
w/ Only object       & 83.7 & 70.2 \\
w/ Only hand         & 83.6 & 67.8 \\
w/ Only task         & 83.2 & 68.4 \\
w/ Object-hand-task  & \textbf{84.9} & \textbf{79.5} \\
\Xhline{1pt}
\end{tabular*}
\vspace{-4mm}
\label{Tab:Ablation_Representation}
\end{wraptable}







       

\textbf{Impact of Structured Triadic Representations.}
In Table~\ref{Tab:Ablation_Representation}, we analyze the impact of different triadic representation components. Using a single representation leads to limited generalization, 
\gh{which indicates} that no single cue alone can support cross-scene, cross-object, and cross-task transfer. 
Jointly using object, hand, and task representations achieves the best results, especially on generalization tasks. This shows that structured triadic representations provide more complete and transferable cues for action prediction.


       
\begin{wraptable}{r}{0.42\textwidth}
\vspace{-7mm}
\centering
\scriptsize
\setlength{\tabcolsep}{2.8pt}
\renewcommand{\arraystretch}{0.85}
\caption{Effectiveness of loss combination.}
\begin{tabular*}{0.42\textwidth}{@{\extracolsep{\fill}}ccc}
\Xhline{1pt}
Loss & Fine-tuned & Generalized \\
\hline
w/ $\mathcal{L}_{act}$ & 84.0 & 74.2 \\
w/ $\mathcal{L}_{act}+\mathcal{L}_{obj}+\mathcal{L}_{hand}$ & \textbf{84.9} & \textbf{79.5} \\
\Xhline{1pt}
\end{tabular*}
\vspace{-4mm}
\label{Tab:Ablation_Losses}
\end{wraptable}

\textbf{Effectiveness of Losses.}
\gh{Table~\ref{Tab:Ablation_Losses} shows that adding object and hand auxiliary constraints to the action loss improves both fine-tuning and generalization by stabilizing object–hand alignment and enhancing relational modeling and action prediction.}

	
       
\begin{wraptable}{r}{0.44\textwidth}
\vspace{-6mm}
\raggedleft
\scriptsize
\renewcommand{\arraystretch}{0.85}
\caption{Effect of proprioception on hand.}
\begin{tabular*}{0.43\textwidth}{@{\extracolsep{\fill}}ccc}
\Xhline{1pt}
Strategy & Fine-tuned & Generalized \\
\hline
w/o Proprio & 83.2 & 71.6 \\
w/ Proprio  & \textbf{84.9} & \textbf{79.5} \\
\Xhline{1pt}
\end{tabular*}
\vspace{-3mm}
\label{Tab:Discussion_Proprio}
\end{wraptable}

\textbf{Effect of Proprioception on Hand Representation.}
\gh{Table~\ref{Tab:Discussion_Proprio} shows that proprio-anchored hand tokens outperform visual-only tokens, especially in generalization, by improving end-effector localization under occlusion and viewpoint changes and strengthening state–visual alignment.}

	
       
\begin{wraptable}{r}{0.40\textwidth}
\vspace{-6mm}
\centering
\scriptsize
\setlength{\tabcolsep}{2pt}
\renewcommand{\arraystretch}{0.85}
\caption{Task representation on nodes.}
\begin{tabular*}{0.40\textwidth}{@{\extracolsep{\fill}}cccc}
\Xhline{1pt}
Object & Hand & Fine-tuned & Generalized \\
\hline
w/o Task-Gnd & w/o Task-Gnd & 84.1 & 77.2 \\
w/ Task-Gnd  & w/o Task-Gnd & 84.3 & 77.8 \\
w/ Task-Gnd  & w/ Task-Gnd  & \textbf{84.9} & \textbf{79.5} \\
\Xhline{1pt}
\end{tabular*}
\vspace{-5mm}
\label{Tab:Discussion_Nodes}
\end{wraptable}

\textbf{Impact of Task Representations on Nodes.}
Table~\ref{Tab:Discussion_Nodes} analyzes the effect of task representations on object and hand node construction. 
\gh{Without task representations, the model performs well on fine-tuned tasks but generalizes poorly as nodes lack task association. Task representations improve both, especially generalization, by guiding nodes into task-driven representations for cross-object and cross-task transfer.}

   		
       
\begin{wraptable}{r}{0.4\textwidth}
\vspace{-6mm}
\raggedleft
\scriptsize
\setlength{\tabcolsep}{2.5pt}
\renewcommand{\arraystretch}{0.85}
\caption{Discussion on bottleneck space.}
\begin{tabular*}{0.4\textwidth}{@{\extracolsep{\fill}}cccc}
\Xhline{1pt}
Setting & Succ$\uparrow$ & GFLOPs$\downarrow$ & Mem (GB)$\downarrow$ \\
\hline
w/o Bottleneck & 79.3 & 742.5 & 24.8 \\
w/ Bottleneck  & \textbf{79.5} & \textbf{618.3} & \textbf{19.6} \\
\Xhline{1pt}
\end{tabular*}
\vspace{-5mm}
\label{Tab:Discussion_Bottleneck}
\end{wraptable}

\textbf{Role of Relational Bottleneck Representation.}
\gh{Table~\ref{Tab:Discussion_Bottleneck} shows relational bottleneck reduces cost without sacrificing performance by compressing relation-enhanced nodes, preserving action-relevant cues, and filtering redundancy.}

\textbf{Limitation.}
\gh{Although} TriRelVLA performs well in cross-scene, cross-object, and cross-task manipulation, 
\gh{it} still struggles with long-horizon compositional and unseen tasks. This is mainly due to the lack of explicit memory-based reasoning over long-term triadic spatiotemporal relations and multi-stage subgoals. 
\gh{Our} future work will introduce long-term memory \cite{lei2025robomemory}, task decomposition \cite{ahn2022can}, and spatiotemporal relational graph modeling for more complex compositional manipulation.

\vspace{-2mm}
\section{Conclusion}
\label{sec:conclusion}
\vspace{-2mm}
In this work, we propose \textbf{TriRelVLA}, a VLA model with a triadic relational structure for generalizable embodied manipulation. We extract object, hand, and task tokens from multimodal inputs as structured cues bridging perception and action generation. A task-guided relational graph models object-hand-task interactions 
\gh{to enable} action decisions based on structured constraints 
\gh{instead of} entangled visual appearance. 
With a compact relational bottleneck, TriRelVLA compresses and aligns triadic relations with linguistic representations in the LLM for robust action prediction. We further construct datasets with object/hand mask annotations and introduce auxiliary supervision for stable relation-driven learning. Experiments demonstrate the superiority of TriRelVLA, especially in generalization to unseen scenes, objects, and tasks.

{
  \small
  \bibliographystyle{unsrt}
  \bibliography{egbib}
}


\end{document}